\documentclass[11pt, a4paper, logo, copyright, nonumbering]{deepseek}
\usepackage[authoryear, sort&compress, round]{natbib}
\usepackage{dblfloatfix}
\usepackage{ulem}
\usepackage{caption}
\usepackage{listings}
\usepackage{verbatim}
\usepackage{dramatist}
\usepackage{xspace}
\usepackage{pifont}
\usepackage{multirow}
\usepackage{xltabular}
\usepackage{longtable}
\usepackage{hyperref}
\usepackage{tcolorbox}
\usepackage{tikz}
\usepackage{fontawesome5} 
\tcbuselibrary{breakable} 
\tcbuselibrary{skins,listings} 
\usepackage{amsfonts}
\usepackage{amsmath}
\usepackage{xcolor} 
\usepackage{amssymb}
\usepackage{lineno}
\usepackage{adjustbox}
\usepackage{graphicx}
\usepackage[bottom]{footmisc}
\usepackage{float}

\usepackage{CJKutf8}
\usepackage{subcaption}
\usepackage{setspace}

\usepackage{dsfont}
\usepackage{array}
\usepackage{tabularx}
\usepackage{booktabs}
\usepackage{lipsum}
\usepackage{multicol}
\definecolor{promptgray}{gray}{0.9} 
\makeatletter
\def\@BTrule[#1]{%
  \ifx\longtable\undefined
    \let\@BTswitch\@BTnormal
  \else\ifx\hline\LT@hline
    \nobreak
    \let\@BTswitch\@BLTrule
  \else
     \let\@BTswitch\@BTnormal
  \fi\fi
  \global\@thisrulewidth=#1\relax
  \ifnum\@thisruleclass=\tw@\vskip\@aboverulesep\else
  \ifnum\@lastruleclass=\z@\vskip\@aboverulesep\else
  \ifnum\@lastruleclass=\@ne\vskip\doublerulesep\fi\fi\fi
  \@BTswitch}
\makeatother

\addto\extrasenglish{

}

\reportnumber{001}

\renewcommand{\today}{}

\title{\centering TBStar-Edit: From Image Editing Pattern Shifting to Consistency Enhancement}

\author[*]{
Hao Fang,\hspace{0.25em} Zechao Zhan,\hspace{0.25em} Weixin Feng,\hspace{0.25em} Ziwei Huang,\hspace{0.25em} Xubin Li,\hspace{0.25em} Tiezheng Ge \\
{ Taobao \& Tmall Group of Alibaba}
}

\renewcommand{\phi}{\varphi}

\renewcommand{\epsilon}{\varepsilon}
\renewcommand{\imath}{\mathrm{i}}

\newlength{\restsubwidth}
\newlength{\restsubheight}
\newlength{\restsubmoreheight}
\newcommand{\rest}[2]{%
        \settowidth{\restsubwidth}{\ensuremath{#2}}
        \settoheight{\restsubheight}{\ensuremath{{}_{#2}}}
        \ensuremath{{#1\hskip 0.5pt}_{\vrule\kern2pt\parbox[b][%
        4pt][b]{\the\restsubwidth}{%
                        \ensuremath{{}_{#2}}}}}
        }

\begin{abstract}

Recent advances in image generation and editing technologies have enabled state-of-the-art models to achieve impressive results in general domains. However, when applied to e-commerce scenarios, these general models often encounter consistency limitations. To address this challenge, we introduce TBStar-Edit, an new image editing model tailored for the e-commerce domain. Specifically, for data engineering, we establish a comprehensive data construction pipeline, encompassing data collection, construction, filtering, and augmentation, to acquire high-quality, instruction-following, and strongly consistent editing data to support model training. For model architecture design, we design a hierarchical model framework consisting of a base model, pattern shifting modules, and consistency enhancement modules. For model training, we adopt a two-stage training strategy to enhance the consistency preservation: first stage for editing pattern shifting, and second stage for consistency enhancement. Each stage involves training different modules with separate datasets. Finally, we conduct extensive evaluations of TBStar-Edit on a self-proposed e-commerce benchmark, and the results demonstrate that TBStar-Edit outperforms existing general-domain editing models in both objective metrics (VIE Score) and subjective user preference.

  \end{abstract}
  
\begin{document}
    \begin{figure}[htbp]
    \centering
    \includegraphics[width=0.85\linewidth]{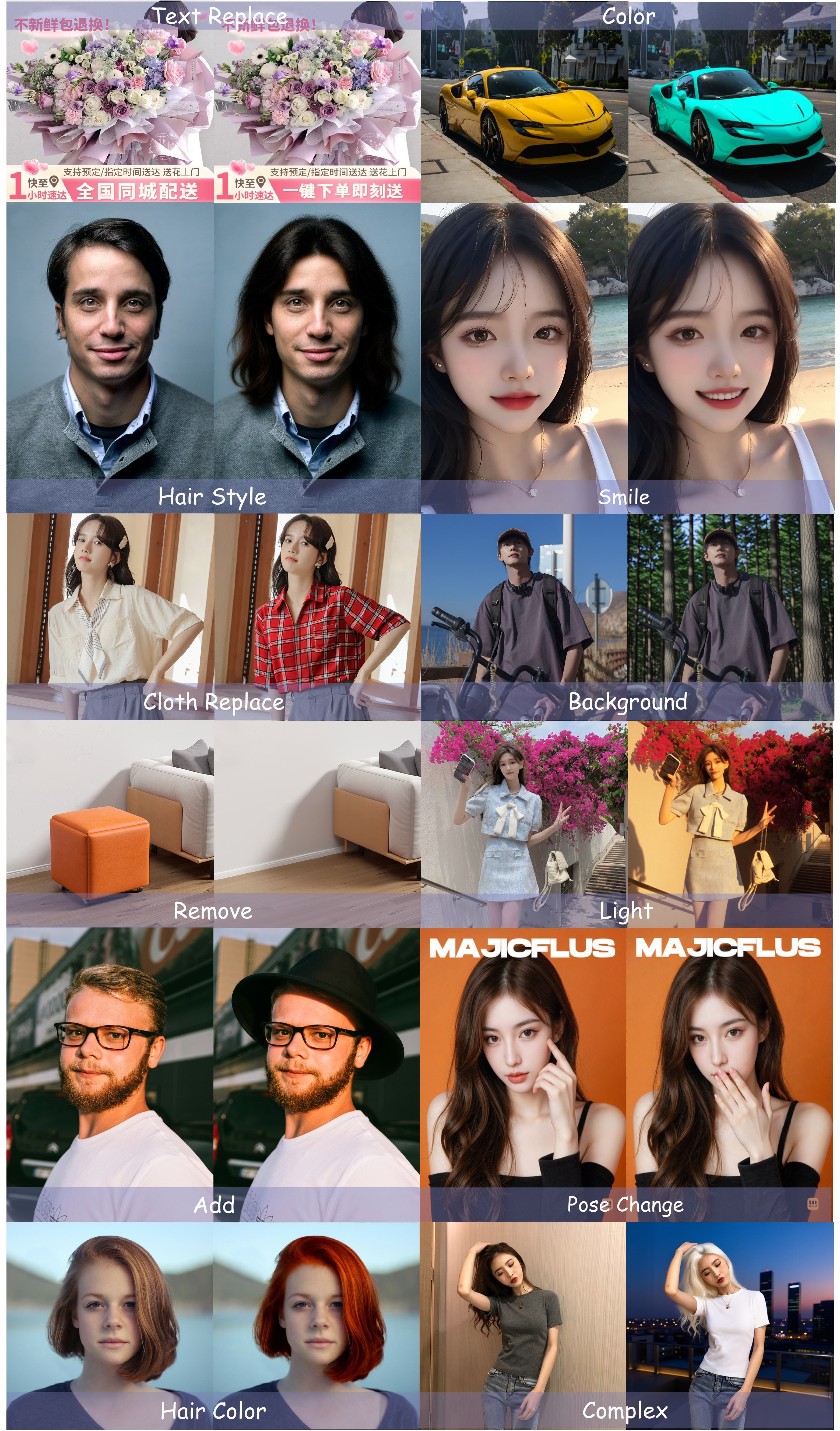} 
    \caption{Overview of TBStar-Edit. TBStar-Edit can handle a variety of tasks in both general and e-commerce domains.}
    \label{fig:gen_images}
  \end{figure}
  \maketitle
  

\section{Introduction}

Driven by the rapid advancements in Diffusion Models \cite{ho2020denoising, lipman2022flow} and text-to-image generation \cite{esser2024scaling, blackforestlabs}, AI-powered image editing technologies have experienced significant progress in recent years. From early UNet-based models such as Instruct-pix2pix \cite{brooks2023instructpix2pix}, EmuEdit \cite{sheynin2024emu}, and HIVE \cite{zhang2024hive} to more recent DiT-based models \cite{peebles2023scalable} like ACE \cite{han2024ace}, ACE++ \cite{mao2025ace++}, and Step1X-Edit \cite{liu2025step1x}, substantial breakthroughs have been made in terms of model parameters, the quantity and quality of training data and the resolution of generated images. Currently, editing models feature increasingly versatile editing capabilities and noticeably improved editing performance. Numerous companies have introduced multimodal editing models, including GPT-4o \cite{hurst2024gpt}, Nano Banana \footnote{\url{https://aistudio.google.com/models/gemini-2-5-flash-image}}, Flux-Kontext \cite{batifol2025flux}, HiDream-E1 \cite{hidreami1technicalreport}, Qwen-Image-Edit \cite{wu2025qwen}, and SeedEdit \cite{wang2025seededit, seedream2025seedream}. These general-domain editing models exhibit powerful abilities in complex semantic understanding, multilingual scenarios, and creative editing, making rapid image modification a practical reality.

Although these models have demonstrated impressive performance in general domains, their performance is still limited in e-commerce scenarios. The primary reason for this limitation is inadequate consistency preservation, which often manifests as unintended changes to the main subject or background when local modifications are made. Consequently, discrepancies may arise in the identity of the product or in fine-grained details between the edited and original images. To address this issue, we propose TBStar-Edit, a model specifically designed for e-commerce scenarios that require high consistency. By employing rigorous data engineering, model architecture, and training strategies, TBStar-Edit enables precise and high-fidelity image editing while maintaining the integrity of the product and its layout.

For data engineering, we observe that existing open-source datasets such as HQEdit \cite{hui2024hq}, UltraEdit \cite{zhao2024ultraedit}, OmniEdit \cite{wei2024omniedit}, and GPT-Image-Edit \cite{wang2025seededit} suffer from limitations including imprecise instruction-following, insufficient consistency preservation, and suboptimal image quality. Relying solely on these open-source datasets makes it challenging for models to learn high-consistency, instruction-driven editing effectively. To address the limitations of existing data, we propose a systematic four-stage pipeline for data construction. Each stage is carefully designed: in data collection, we prioritize real-world images; in editing pair construction, we use four specialized methods for different tasks; during data filtering, we combine automated vision-language models \cite{Qwen-VL, bai2025qwen2, qwen3max, comanici2025gemini} with manual review to ensure quality; and in post-processing, we diversify the editing instructions through large language models \cite{achiam2023gpt, comanici2025gemini, yang2025qwen3}. This pipeline ensures that our final dataset features precise instruction-following, strong subject consistency, and high image quality, providing solid support for consistency enhancement.

For model architecture, we design a hierarchical model architecture featuring a base model with two plug-and-play components: pattern shifting modules and consistency enhancement modules. This layered structure enables our model to flexibly adapt to various backbone models and aligns with our training strategy. For training strategy, we divide the process into two stages. The first stage establishes initial editing capabilities by training both the pattern shifting and consistency enhancement modules on a mix of open-source and our constructed data. The second stage then enhances consistency by training the consistency enhancement module solely with a high-quality, self-constructed dataset.

Finally, we conduct a comprehensive evaluation on our self-constructed e-commerce benchmark, EcomEdit-Bench. We compare TBStar-Edit against state-of-the-art (SOTA) general-domain editing models, including Flux-Kontext, Qwen-Image-Edit, NanoBanana, and Seedream 4.0 \cite{seedream2025seedream}. The assessment covered three key aspects: instruction-following, consistency preservation, and image quality. The results, derived from visualization, VIE Score \cite{ku2023viescore} calculations, and a user preference study, confirm the superiority of our model. Notably, in the user preference study, TBStar-Edit was favored over the SOTA models by a significant margin of approximately 11$\%$.

\begin{figure}[ht]
  \centering
  \includegraphics[width=1\linewidth]{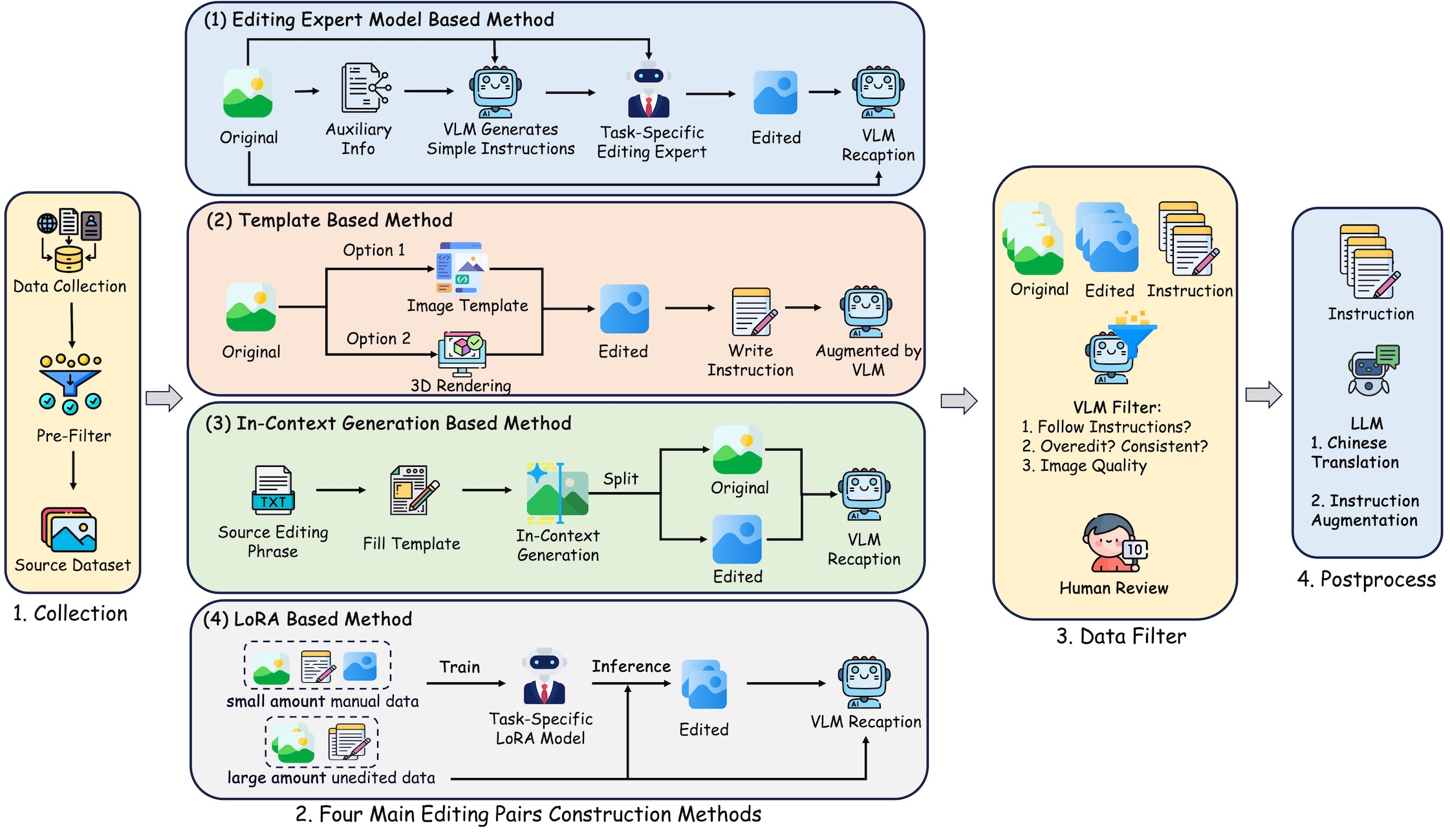} 
  \caption{Pipeline of data engineering.}
  \label{fig:datapipeline}
\end{figure}

\section{Data Engineering}

The performance of an image editing model is critically dependent on the quality of its training data. To enable a model to learn precise, high-fidelity modifications, the data must exhibit three key attributes: accurate instruction following, strong subject consistency, and high image quality. However, existing open-source datasets often suffer from significant limitations, such as severe detail distortion and identity (ID) drift. To overcome these challenges, we have developed a comprehensive data pipeline. This pipeline is structured into four stages: data collection, editing pair construction, data filtering, and data post-processing, as illustrated in Figure \ref{fig:datapipeline}.

\noindent
\subsection{Data Collection}
Recent paradigms in instruction-based image editing often leverage large-scale synthetic datasets for training. While this approach significantly reduces the difficulty of data acquisition and enhances data diversity, an over-reliance on synthetic data can diminish the realism of editing results, hindering a model's generalization and fidelity in real-world scenarios. To address this, we propose a data collection pipeline to collect a large number of high-fidelity, real-world images. To ensure the diversity of our source data, we gathered images from a variety of scenes and sources, including general-domain portraits and object images from the web \footnote{\url{https://unsplash.com}} \footnote{\url{https://pixels.com}} and open-source datasets \cite{ye2025imgedit, wei2024omniedit}, as well as internal e-commerce portraits and product images. Lastly, to guarantee the quality of the source images, we employ automated metrics (e.g., image resolution and Aesthetic Scores) for filtering. 
\noindent
\subsection{Editing Pairs Construction Methods}

The construction of editing pairs from raw image data is the fundamental step in creating image editing datasets. Image editing tasks are highly diverse, each with unique properties and requirements. This diversity necessitates tailored approaches for data generation. Accordingly, we develop four distinct methods for pair construction:  the Editing Expert Model Based, the Template Based, the In-Context Generation Based, and the LoRA \cite{hu2022lora} Based Method.

\begin{figure}[ht]
  \centering
  \includegraphics[width=1\linewidth]{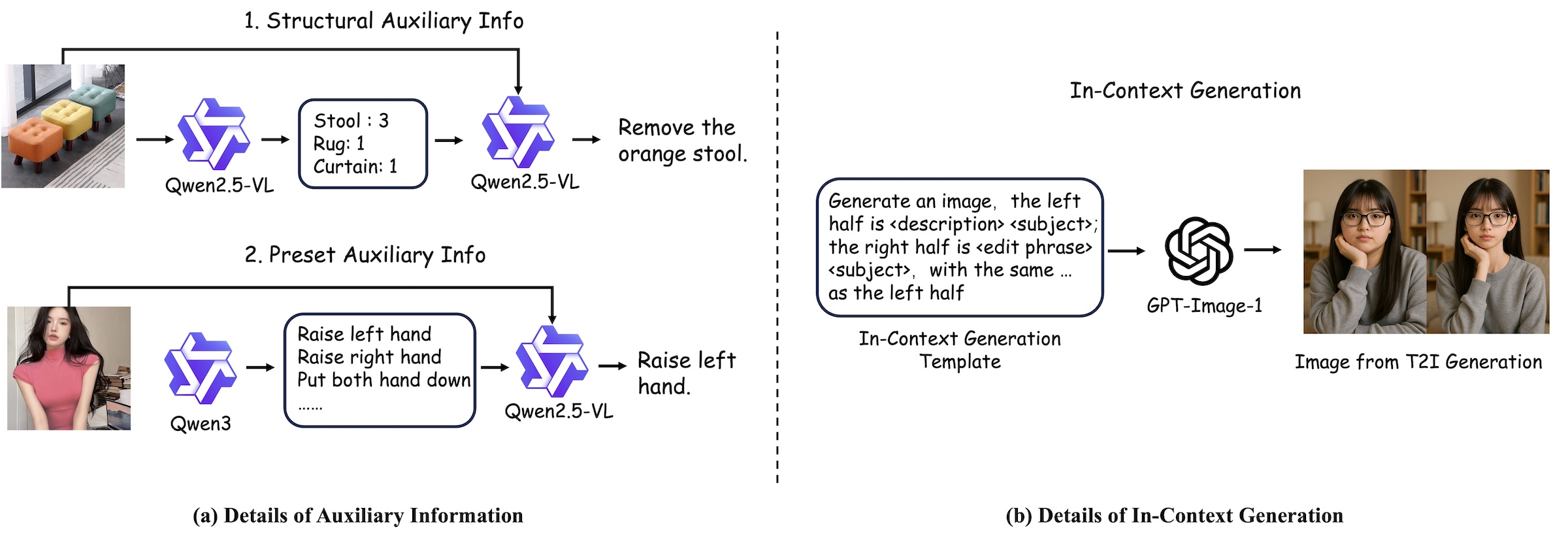} 
  \caption{Details of Auxiliary Information and In-context Generation in editing pairs construction.}
  \label{fig:detail}
\end{figure}

\subsubsection{Editing Expert Model Based Method} 
Although numerous models have demonstrated exceptional performance on various editing tasks, their applicability is often restricted to specific domains. To harness the specialized strengths of these models and integrate their capabilities, we introduce a novel pipeline that employs a Vision-Language Model (VLM) for instruction generation and Task-specific Editing Experts for image manipulation.

Conventional approaches involve feeding raw images directly into a VLM to generate instructions. This method, however, is susceptible to the VLM's inherent biases, which can lead to a lack of diversity and poor generalization in the generated instructions. For instance, a VLM might exhibit a bias towards removing small, peripheral objects while overlooking more prominent subjects in the image. To mitigate these biases, we augment the VLM's input with two types of auxiliary information alongside the original image: (1) a list of object categories with their respective counts, and (2) a set of pre-defined instructions, as detailed in Figure \ref{fig:detail}(a). By conditioning the VLM on this enriched input, we aim to reduce its biases and thereby enhance the diversity and generalizability of the final instruction set.

The VLM-generated instructions and the original images are then processed by a Task-specific Editing Expert (e.g., an open-source model, a proprietary workflow) to create the edited images. In the final stage, this new image pair (original and edited) is used to prompt the VLM again, this time to generate a more detailed and descriptive instruction. This methodology strategically leverages auxiliary data and specialized experts to facilitate the rapid generation of a large-scale, diverse dataset,  which boosts performance on targeted editing tasks.

\subsubsection{Template Based Method}
For certain editing tasks that can be templated, such as watermark removal, text element removal, text modification, and lighting/shadow adjustment, we employ a template-based data construction approach. The main workflow of this method involves: (1) generating edited images by adding image templates to the original image, applying 3D rendering templates, or modifying specific template elements in relevant regions of the original image; (2) generating editing instructions either through predefined rules or manual input, followed by enhancement of instruction diversity using a VLM. As this approach ensures that regions outside the templates remain completely unchanged, the resulting data exhibits excellent subject consistency.

\subsubsection{In-Context Generation Based Method}
Although the expert-based and template-based methods address most editing needs, they fall short for tasks that lack corresponding experts or cannot be easily templated, such as modifying a person's age or physique. For these cases, we adopt an in-context generation-based method. This approach, inspired by the in-context learning abilities demonstrated in works like IC-LoRA \cite{huang2024context}, ICEdit \cite{zhang2025context} and UNO \cite{wu2025less}, fully utilizes T2I models to generate identity-preserving image pairs or sequences (see Figure \ref{fig:detail}(b)). Subsequently, these outputs are split, and a VLM generates the editing instructions. Due to its flexibility, this method is particularly well-suited as a supplementary data construction strategy for niche editing tasks.

\subsubsection{LoRA Based Method}
For specialized editing tasks that lack editing experts and cannot be templated, we employ an alternative approach by utilizing LoRA for less-to-more data generation. Initially, a small number of editing pairs (ranging from several dozen to a few hundred) are created for these specific tasks, either through web scraping or manual annotation. These pairs are then used to train a task-specific LoRA model. Once LoRA training is complete, we input a large set of raw image-instruction pairs—annotated by VLM—into the task-specific LoRA model, which infers and produces a substantial volume of edited images. Subsequently, both the original and edited images are input into the VLM for refined instruction generation. Compared to the In-Context Generation Based Method, this approach better ensures the authenticity of the data and helps improve inference performance in real-world scenarios, although it demands more time and human resources.

\noindent
\subsection{Data Filter and Data Postprocess}
Although our editing data construction methods are rigorously designed to ensure high quality and a high acceptance rate for edited images, limitations inherent to various editing models and persistent biases in VLMs cannot be entirely eliminated. As a result, a certain proportion of low-quality edited data inevitably remains in our constructed dataset, potentially affecting model training performance. To further guarantee the quality of the editing data, we implement a strict filtering process for the constructed editing pairs. First, all constructed data are evaluated by a VLM, which scores them across multiple dimensions, including instruction-following, consistency between the original and edited images, and image quality. Low-scoring data are excluded based on these assessments. Furthermore, due to discrepancies between VLM and human judgment regarding the quality of edited data, we conduct manual quality inspection to obtain a set of high-quality data.

Finally, to ensure the model responds precisely to multilingual instructions (e.g., Chinese and English), we augment the editing instructions in two ways. First, the original instructions are translated into Chinese. Second, a large language model is used to generate diverse phrasings for both languages, including synonyms, interrogative forms, and passive constructions. The complete collection of original and augmented instructions is then compiled for flexible use during model training.

\begin{figure}[ht]
    \centering
    \includegraphics[width=0.8\linewidth]{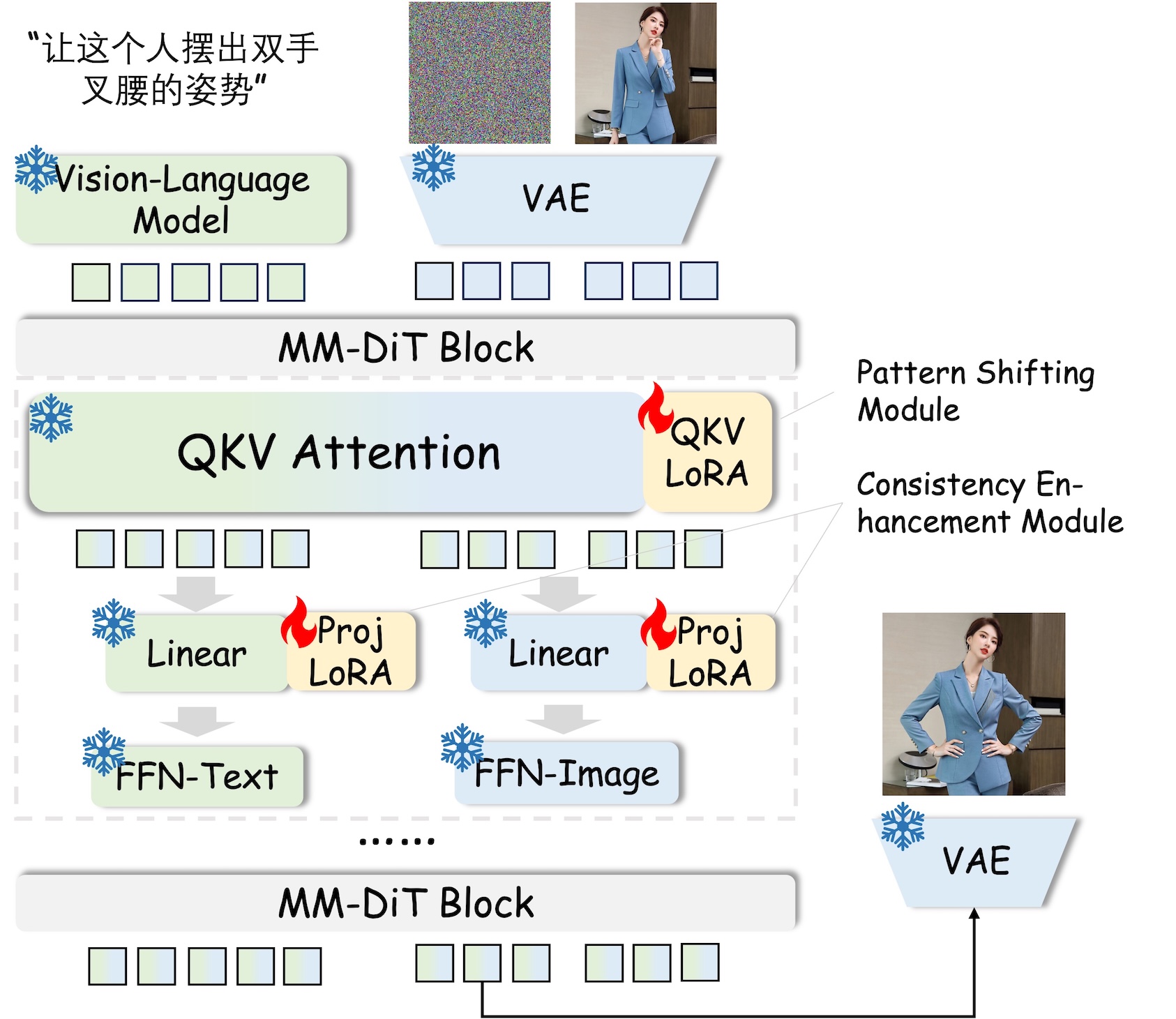} 
    \caption{Framework of the TBStar-Edit model.}
    \label{fig:model}
\end{figure}

\section{TBStar-Edit Model}
\label{sec:model_architecture}

TBStar-Edit is built upon two key principles: (1) fully leveraging the powerful in-context generation capability of T2I model and transforming it into image editing capability; (2) focusing the capability of the converted editing model to enhance subject consistency in edited images, ensuring practical deployment in real-world scenarios. To cover these principles, we design a flexible, hierarchical model framework that can be adapted to various T2I backbone models, along with a training strategy that first transforms a T2I generation model into an image editing model, and subsequently enhance consistency within the shifted editing model.

\subsection{Model Architecture}
\label{subsec:text_encoder}

Our model utilizes a hierarchical framework composed of a base model, pattern shifting modules, and consistency enhancement modules, a design that affords high flexibility and scalability. This architecture facilitates seamless adaptation to various open-source or proprietary backbone models. Specifically, within each DiT-Block, the pattern shifting module is implemented by inserting a single LoRA layer at the QKV Attention position \cite{vaswani2017attention}. The consistency enhancement module, in turn, is composed of two additional LoRA layers applied to the Projections following the Attention block. During training, we freeze all parameters of the original backbone model and optimize only those of the pattern shifting and consistency enhancement modules. We evaluate our framework's editing adaptation and capability on several models, including Flux.1-Dev, Flux.1-Kontext-Dev, HiDream-l1, and Qwen-Image. Among these, Qwen-Image demonstrated exceptional performance in Chinese contexts, particularly in generating Chinese e-commerce text and parsing complex instructions. Consequently, we select Qwen-Image as the base model for TBStar-Edit.

\subsection{Training Strategy}
\label{subsec:dit}

\paragraph{Multi-stage Training Strategy.}
Our training process has two distinct phases: first to establish foundational editing capabilities, and second to refine subject consistency. \textbf{Stage 1: Pattern Shifting.} The initial phase aims to equip the model with broad-spectrum editing capabilities. To achieve this, we train the model on a composite dataset combining large-scale open-source data with our high-quality proprietary data. During this stage, all LoRA parameters are made trainable, encompassing the pattern shifting and consistency enhancement modules. \textbf{Stage 2: Consistency Enhancement.} The second phase specializes the model for key e-commerce editing tasks and addresses residual consistency issues from the first stage. To preserve the general editing abilities acquired previously, training is conducted exclusively on our high-quality proprietary datasets. The parameters of the pattern shifting modules are frozen, and only the consistency enhancement modules are optimized.

\paragraph{Mixed Chinese-English Instruction Training.}
To improve the model's response precision and editing consistency, especially for Chinese instructions, we train the model on a mixed dataset. This dataset comprises both original and augmented instructions in Chinese and English, sampled at a specific ratio.


\begin{figure}[htbp]
  \centering
  \includegraphics[width=0.9\linewidth]{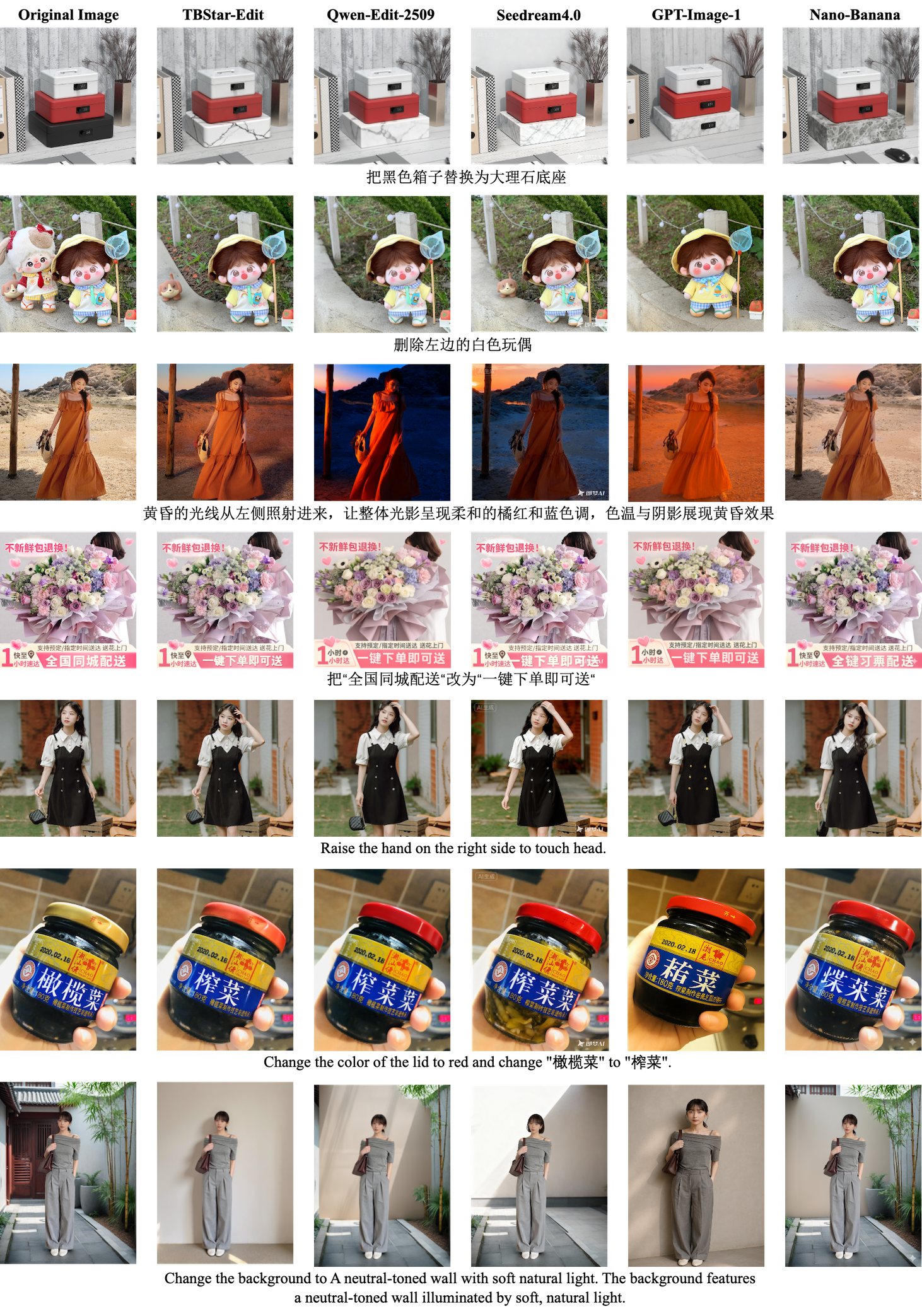} 
  \caption{visualization comparison between state-of-the-art models in general domain.}
  \label{fig:res}
\end{figure}

\section{Experiment}
\label{sec:training}

\subsection{Visual Comparison}
\label{subsec:pretraining}

The visualization results for several tasks are shown in Figure \ref{fig:res}. It can be observed that TBStar-Edit achieves strong performance not only on general tasks such as object replacement and object removal, but also on e-commerce-specific scenarios including portrait background replacement and lighting adjustment. The edited images demonstrate a high degree of balance between instruction following and subject consistency. Compared with other models, TBStar-Edit excels particularly in consistency. While maintaining faithful adherence to the input instructions, it delivers best-in-class results on identity preservation, action retention, and detail maintenance in portrait editing among all evaluated models.

\subsection{Quantitative Evaluation}
\label{subsec:posttraining}

To assess model performance on e-commerce editing tasks, we developed EcomEdit-Bench, a new evaluation benchmark inspired by the widely-adopted, general-domain GEdit-Bench \cite{liu2025step1x}. This benchmark comprises about 380 image-instruction pairs covering a diverse range of common e-commerce scenarios. These include portrait modifications (e.g., pose editing, background replacement, hairstyle changes), object manipulation (e.g., adding accessories, removing or replacing elements), and image-level adjustments such as text editing, watermark removal, and lighting correction.

For evaluation, we employ an automated and objective protocol. Each generated image is scored by GPT-4o across three key standards, including \textbf{Instruction Adherence} (How well the image follows the text prompt), \textbf{Subject Consistency} (The degree to which the core subject is preserved),
and \textbf{Image Quality} (The overall visual fidelity and realism of the output). These individual scores are then processed to calculate overall VIE Score.

We evaluate the e-commerce editing performance of several prominent models on EcomEdit-Bench, including Flux-Kontext-Dev, Qwen-Image-Edit-2509, NanoBanana, and Seedream4.0. The first two models are evaluated via local deployment of their open-source code, while the latter two are assessed through API calls. Notably, some test cases for the API-based models are excluded due to their internal security filters. GPT-Image-1 is omitted from this quantitative analysis due to limitations in its output resolution and poor consistency. The objective evaluation results, measured by the VIE Score, are presented in Table \ref{tab:vie}. As the table indicates, TBStar-Edit achieves the highest overall score and surpasses all other models in Semantic Consistency. These findings demonstrate that TBStar-Edit strikes an effective balance between instruction adherence and subject consistency, making it highly suitable for applications like e-commerce that demand robust subject preservation.

\begin{table}[t]
  \centering
  \setlength{\tabcolsep}{10pt}
  \renewcommand{\arraystretch}{1.15}
  \caption{Evaluation results on EcomEdit-Bench. $G_{SC}$: Semantic Consistency, $G_{PQ}$: Perceptual Quality, $G_O$: Overall. Bold font indicates the highest value.}
  \label{tab:vie}
  \begin{tabular}{lccc}
    \toprule
    Model           & $G_{SC} \uparrow$ & $G_{PQ} \uparrow$ & $G_O \uparrow$ \\
    \midrule
    Flux-Kontext-Dev    & 7.552   & 8.441  & 7.732  \\
    HiDream-E1          & 6.849   & 7.701  & 6.658   \\
    SeedEditv3.0-en     & 8.701   & 8.567  & 8.376  \\
    SeedEditv3.0-cn     & 8.757   & 8.595  & 8.410  \\
    NanoBanana          & 8.217   & \textbf{8.775}  & 8.087   \\
    Qwen-Image-Edit-2509-en & 8.971   & 8.617  & 8.579   \\
    Qwen-Image-Edit-2509-cn  & 8.954   & 8.630  & 8.569 \\
    Seedream4.0-en      & 9.002   & 8.701  & 8.643  \\
    Seedream4.0-cn      & 9.041   & 8.756  & 8.714  \\
    \textbf{TBStar-Edit-en} & 9.139 & 8.575  & 8.688   \\
    \textbf{TBStar-Edit-cn} & \textbf{9.206} & 8.579  & \textbf{8.746}   \\
    \bottomrule
  \end{tabular}
\end{table}

\begin{table}[t]
  \centering
  \setlength{\tabcolsep}{10pt}
  \renewcommand{\arraystretch}{1.15}
  \caption{Human preference evaluation.}
  \label{tab:human}
  \begin{tabular}{lccc}
    \toprule
    Model           & TBStar-Edit win & TBStar-Edit loss & Tie\\
    \midrule
    TBStar-Edit vs Flux-Kontext-Dev      & 55.80\%   & 18.90\%   & 25.30\%   \\
    TBStar-Edit vs Qwen-Image-Edit-2509      & 46.94\%    & 26.53\%   & 28.57\%   \\
    TBStar-Edit vs NanoBanana & 44.20\%  & 32.50\%   & 23.40\%    \\
    TBStar-Edit vs Seedream4.0 & 40.00\%  & 28.89\%  & 31.11\%    \\
    \bottomrule
  \end{tabular}
\end{table}


While the VIE Score offers a comprehensive assessment of instruction adherence, image quality, and consistency, its reliance on VLM introduces two inherent limitations: (1) VLMs often struggle to discern subtle features, leading to less precise consistency assessments; and (2) they are prone to hallucinations, which can result in inaccurate judgments of instruction following. Consequently, relying solely on automated metrics is insufficient for a complete evaluation. To complement the automated scoring, we conduct a human evaluation using a pairwise comparison format. For each trial, participants are shown the original image, the corresponding instructions (in both English and Chinese), and two edited images. One image is generated by TBStar-Edit, and the other by a randomly selected competitor from the set of {Flux-Kontext, Qwen-Image-Edit-2509, Seedream4.0, NanoBanana}. The display order of the two images is randomized to prevent bias. Participants then select the superior image based on three criteria: instruction adherence, image quality, and consistency preservation. As shown in Table \ref{tab:human}, the results indicate a strong preference for TBStar-Edit over state-of-the-art models like NanoBanana and Seedream4.0. This higher preference rate underscores its competitive advantage in e-commerce applications, where high subject consistency is paramount.

\section{Conclusion}
\label{sec:conclusion}
In this report, we introduce TBStar-Edit, a novel model designed to achieve high product-level consistency in e-commerce image editing. To overcome the limitations of low-quality open-source data, we develop a comprehensive data pipeline that yields a large-scale, high-quality dataset characterized by precise instruction adherence and strong subject preservation. Guided by the principle of "first establish editing capability then enhance consistency“, we design a hierarchical framework composed of a base model, pattern shifting modules, and consistency enhancement modules. This architecture is trained with a two-stage strategy that first equips the model with general editing abilities and then specifically refines its consistency. Experimental results demonstrate that TBStar-Edit significantly outperforms current state-of-the-art general-domain models, particularly in high-consistency e-commerce scenarios.

\bibliography{main}

\newpage
\section*{Appendix}
\section{More Visualization Results}

\begin{figure}[htbp]
  \centering
  \includegraphics[width=1\linewidth]{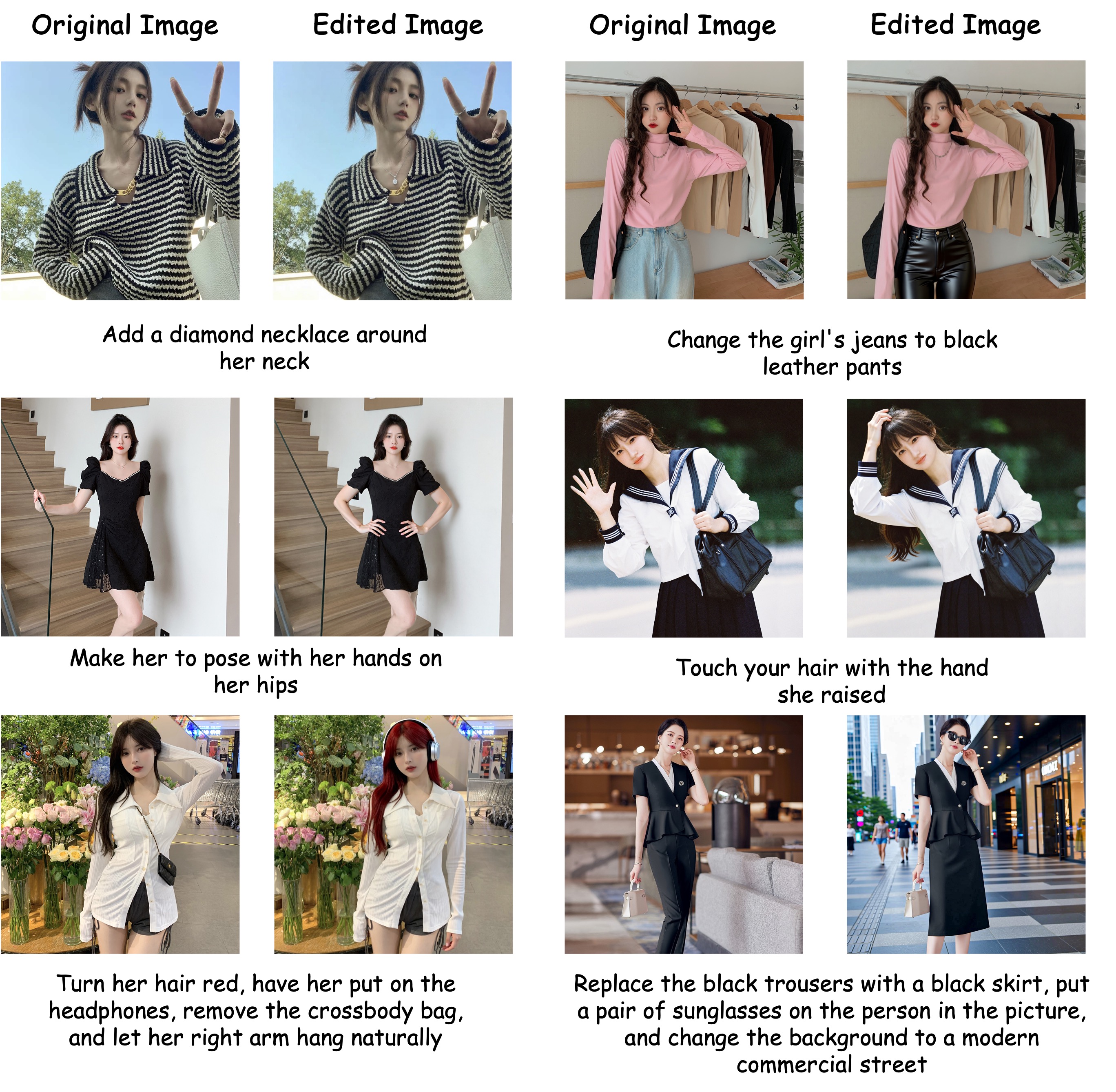} 
  \caption{More visualization results of TBStar-Edit.}
  \label{fig:ap3}
\end{figure}

\begin{figure}[htbp]
  \centering
  \includegraphics[width=1\linewidth]{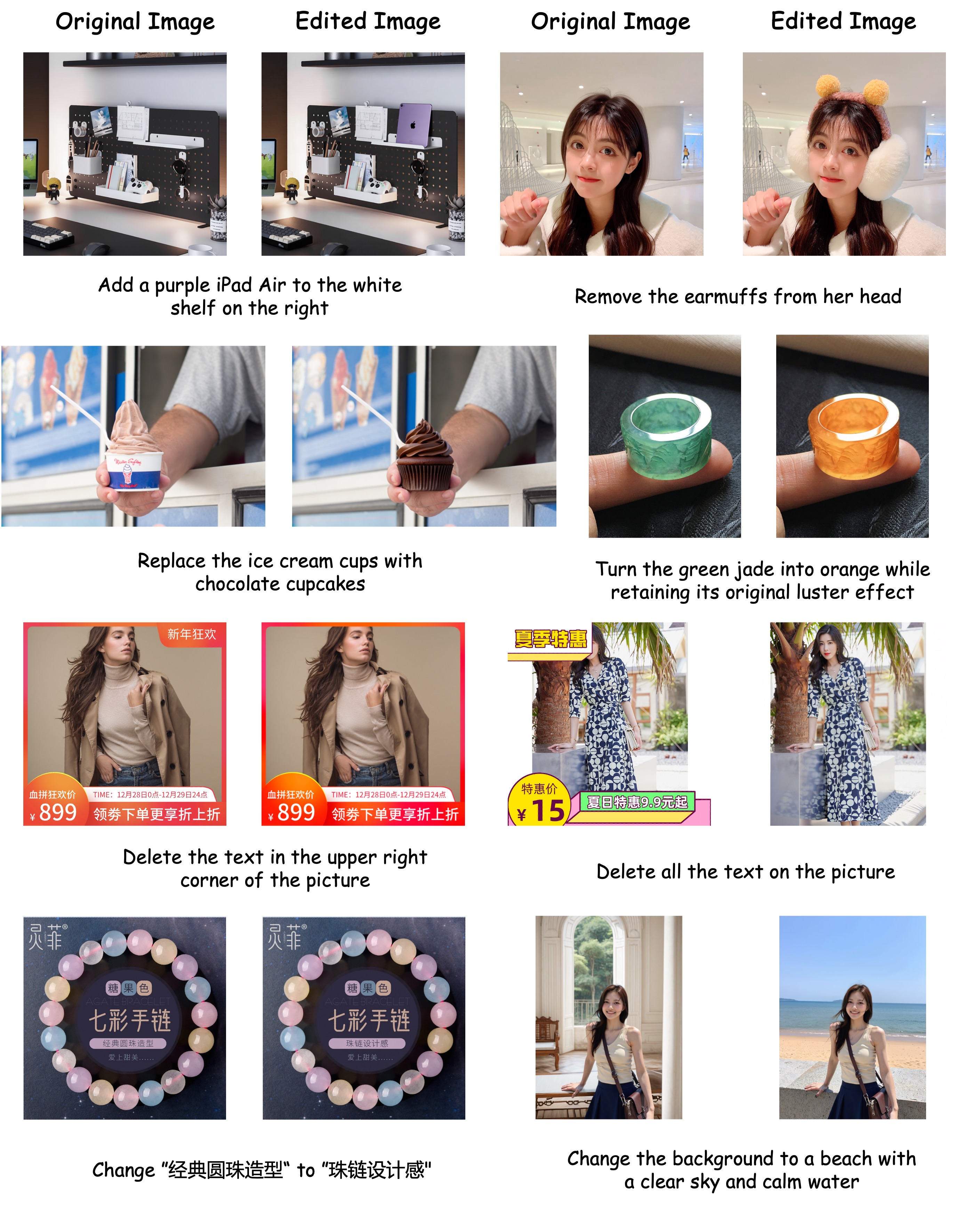} 
  \caption{More visualization results of TBStar-Edit.}
  \label{fig:ap1}
\end{figure}

\begin{figure}[htbp]
  \centering
  \includegraphics[width=1\linewidth]{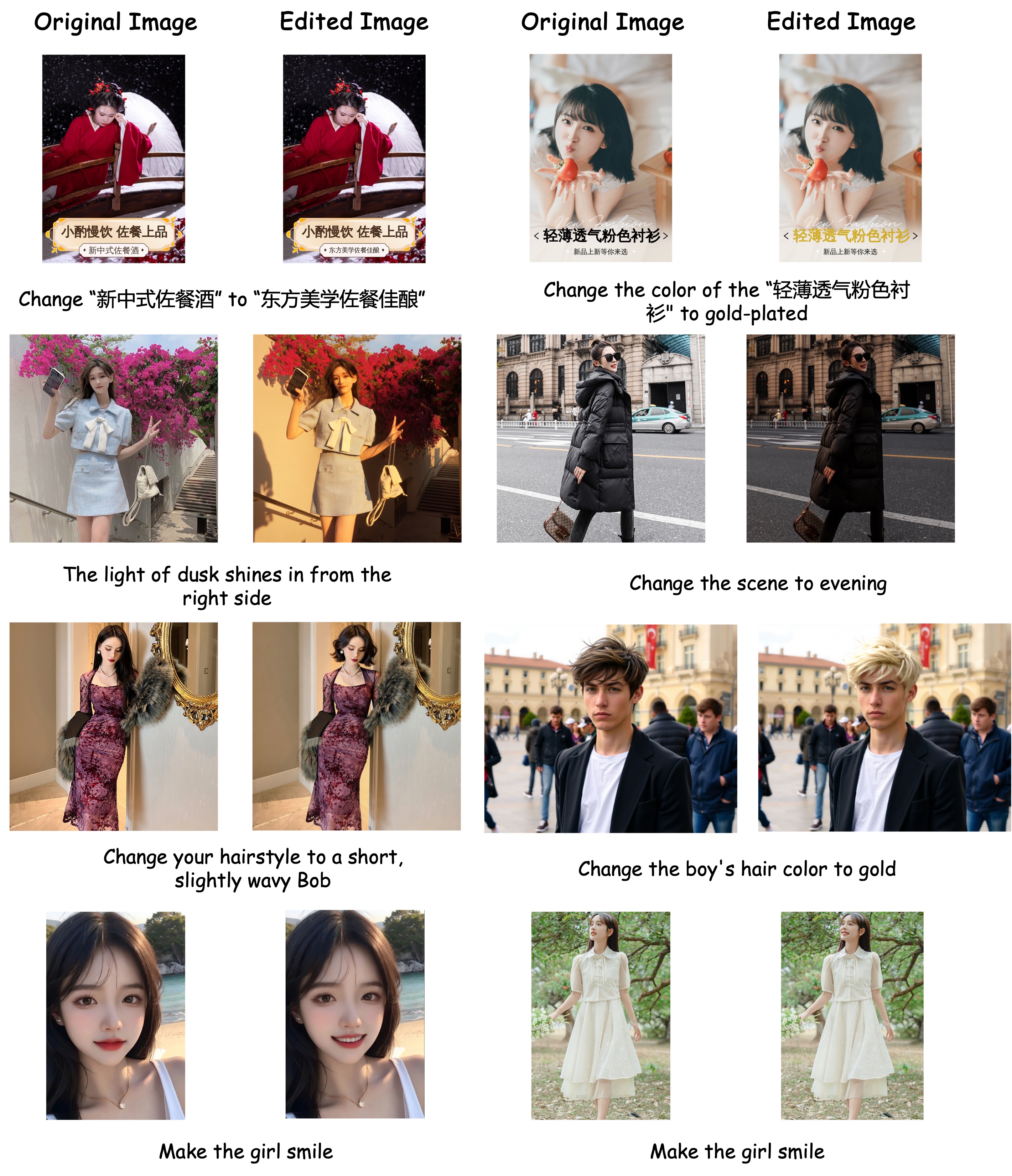} 
  \caption{More visualization results of TBStar-Edit.}
  \label{fig:ap2}
\end{figure}

\setcounter{figure}{0}
\makeatletter 
\renewcommand{\thefigure}{A\@arabic\c@figure}
\makeatother

\setcounter{table}{0}
\makeatletter 
\renewcommand{\thetable}{A\@arabic\c@table}
\makeatother

\end{document}